# Generating Explanations for Evidential Reasoning

**Hong XU**
IRIDIA and Service d'Automatique
Université libre de Bruxelles
50 Ave. F. Roosevelt, CP 194/6
1050-Brussels, Belgium

**Philippe SMETS**
IRIDIA
Université libre de Bruxelles,
50 Ave. F. Roosevelt, CP 194/6
1050-Brussels, Belgium

## Abstract

In this paper, we present two methods to provide explanations for reasoning with belief functions in the valuation-based systems. One approach, inspired by Strat's method, is based on sensitivity analysis, but its computation is simpler thus easier to implement than Strat's. The other one is to examine the impact of evidence on the conclusion based on the measure of the information content in the evidence. We show the property of additivity for the pieces of evidence that are conditional independent within the context of the valuation-based systems. We will give an example to show how these approaches are applied in an evidential network.

## 1 Introduction

The developers of expert systems have realized that a good facility to explain the computer-based reasoning to users is a prerequisite to their more widespread acceptance. The importance of explanation is due to two reasons. First, expert systems are usually used to solve difficult problems. A good explanation facility allows users to observe the inference process that leads the conclusions thereby increases their confidence in the system. Second, an explanation facility helps knowledge engineers refine the problem solving knowledge.

Recently, much attention has been paid to the generation of comprehensible explanations for uncertain reasoning, especially for probabilistic reasoning. One approach is to use sensitivity analysis since it can tell which parameters are most important and most affect the result of the influence. For reasoning with belief functions, Strat [?] has presented some strategies for generating explanation based on sensitivity analysis. Here, we use this idea for generating explanation in a more general network - the valuation-based systems, and show that the computation can be simplified, therefore the implementation is easier.

Another approach for the explanation is to examine the individual impact of evidence on the overall conclusion. A classic technique for probabilistic reasoning is the weight of evidence. Weights of evidence have a useful property of additivity provided that the pieces of evidence are conditional independent. Another term for measuring the impact of a piece of evidence is the amount of information provided by the evidence [15]. Good [2] gave the analysis of these two measures and the relation between them. Within the context of belief functions, Smets [8] has defined a measure of the information content provided by a piece of evidence to show its impact on a frame of discernment. The measure also has the additivity property for the distinct pieces of evidence. In this paper, we use this concept to examine the importance of different pieces of evidence on the hypothesis and show the impact of the pieces of evidence which are conditional independent.

The rest of this paper is as follows: In section 2, we briefly review the basic concepts of evidential reasoning in the valuation-based systems. In section 3, we present the approaches for the explanation through a simple example. In section 4, we discuss the implementation issues of the proposed approaches. Finally in section 5, we give some conclusions.

## 2 Evidential Reasoning

Dempster-Shafer theory [3, 9, 13] is regarded as a useful tool for representing and manipulating uncertain knowledge. It provides flexible input requirements and an efficient method for combining information obtained from multiple sources. In this section, we briefly review reasoning with belief functions in the valuation-based systems. More details can be found in [6].

### 2.1 Basic Concepts

**Definition 1** *Let $\Omega$ be a finite non-empty set called the frame of discernment (the frame for short). The mapping bel: $2^\Omega \to [0, 1]$ is an (unnormalized) belief function if and only if there exists a basic belief assignment (bba) m: $2^\Omega \to [0, 1]$ such that:*

(i) $\sum_{A \subseteq \Omega} m(A) = 1$,

(ii) $bel(A) = \sum_{B \subseteq A, B \neq \emptyset} m(B)$,

(iii) $bel(\emptyset) = 0$.

Those subsets $A$ such that $m(A) > 0$ are called *the focal elements*. A *vacuous belief function* is a belief function such that $m(\Omega)=1$ and $m(A)=0$ for all $A \neq \Omega$, which represents total ignorance.

Given a belief function, we can define a *plausibility function* $pl$: $2^\Omega \to [0, 1]$ and a *commonality function* $q$: $2^\Omega \to [0, 1]$ as follows: for $A \subseteq \Omega$,

$$pl(A) = bel(\Omega) - bel(\bar{A}) \text{ and } pl(\emptyset) = 0$$
$$q(A) = \sum_{A \subseteq B \subseteq \Omega} m(B)$$

where $\bar{A}$ is the complement of $A$ relative to $\Omega$.

Consider two distinct pieces of evidence on $\Omega$ represented by $m_1$ and $m_2$. The belief function $m_{12} = m_1 \oplus m_2$ that quantifies the combined impact of these two pieces of evidence is obtained by the *(unnormalized) Dempster's rule of combination*. The computation is as follows: $\forall A \subseteq \Omega$,

$$(m_1 \oplus m_2)(A) = \sum_{B \cap C = A} m_1(B) m_2(C)$$

If the commonality functions are used, then

$$(q_1 \oplus q_2)(A) = q_1(A) q_2(A).$$

The $m_{12}(\emptyset)$ measures how much $m_1$ and $m_2$ are conflicting [11]. And $k = 1 - m_{12}(\emptyset)$ is a normalization factor in Dempster's rule for getting a normalized belief function. In this paper, we will discard the normalization factor[1] for the computation.

### 2.2 Valuation-Based Systems

Valuation-based systems (VBS) is an abstract framework proposed by Shenoy [6] for uncertainty representation and reasoning. It can represent uncertain knowledge in different domains including probability theory, belief function theory, and possibility theory, etc.. The graphical representation of VBS is called *a valuation network*. A VBS representation consists of a set of variables, and a set of valuations defined on the subsets of variables. The set of all the variables, denoted by **U**, represents the universe of discourse of the problem. For each variable $X_i$, we use $\Theta_{X_i}$ to denote the set of its possible values, and call it *the frame of $X_i$*. For some subset $A(|A| > 1)$ of **U**, a set of valuations defined on $\Theta_A$ represents the relationship among the variables in $A$, where the frame $\Theta_A$ is the Cartesian product of all $\Theta_{X_i}$ for $X_i$ in $A$. We call the knowledge represented by this kind of valuations *the generic knowledge*. In VBS, we can also define the valuations on single variables, which represent the so-called *factual knowledge*. We use **H** to denote the set of all subsets on which the valuations are defined. The valuations are specialized as belief functions in the case for Dempster-Shafer theory. We call such VBS *an evidential reasoning system* or simply *an evidential system*, and the valuation network *an evidential network*.

The goal of evidential reasoning is to assess a certain hypothesis when certain pieces of evidence (factual knowledge) are given. The way to assess the hypothesis is to infer its belief value from the belief values of the evidence. This can be done by evaluating the valuation network by two steps: (1) combine all belief functions in the network, resulting in the so-called *global belief function*; (2) marginalize the global belief function to the frame of each variable or subsets of variables, obtaining *the marginals for each variables or for subsets of variables*. The operations for the reasoning are combination and marginalization which are defined as follows:

*Combination* $\oplus$: Suppose $m_A$ and $m_B$ are two bba's on the subsets $A$ and $B$, then $m_A \oplus m_B$ will be the bba on $A \cup B$ computed by: $\forall c \subseteq \Theta_{A \cup B}$,

$$(m_A \oplus m_B)(c) = \sum_{a^{\uparrow A \cup B} \cap b^{\uparrow A \cup B} = c} m_A(a) m_B(b)$$

where $a^{\uparrow A \cup B}$ and $b^{\uparrow A \cup B}$ are the cylindric extension of $a(\subseteq \Theta_A)$ and $b(\subseteq \Theta_B)$ to $\Theta_{A \cup B}$, respectively.

*Marginalization* $\downarrow$: Suppose $m_A$ is a bba on a subset $A$ and suppose $B \subseteq A$, $B \neq \emptyset$. The marginal of $m_A$ for $B$, denoted by $m^{\downarrow B}$, is a bba on $B$ computed by: $\forall b \subseteq \Theta_B$,

$$m^{\downarrow B}(b) = \sum_{a \subseteq \Theta_A, a^{\downarrow B} = b} m_A(a)$$

where $a^{\downarrow B}$ is a projection of $a$ to $\Theta_B$ by projecting each element of $a$ to $\Theta_B$.

---

[1] The normalization factor has been criticized by Zadeh [19] with a counter-example which shown the danger of its blind application. More discussion about the normalization problem can be found in [9].

The marginal for a variable $X_i$ is computed by: $(\oplus\{bel_B | B \in \mathbf{H}\})^{\downarrow X_i}$.

As it is not feasible to compute the global belief function when there are a large number of variables in the network, Shenoy and Shafer [5] has proposed a local computation technique to compute the marginals for variables without computing the global belief function explicitly. For the details of the technique, readers can refer to [4, 5].

## 3 Explanation of Reasoning Process

One major goal of the work on explanation is to understand the reasoning process of an evidential system. This helps the builders and the users to maintain the system and to use it effectively. Explanations can usually be performed by answering the questions such as: why a specific hypothesis is strongly supported, or not? Which evidence is more influential to the conclusion? etc.. In this section, we present two methods for the explanations by answering such kinds of questions through a simple example[2].

**Example:** The Captain of a ship would like to know how many days late a ship will arrive in port. The goal is to find the *Arrival delay*, or by how many days the ship will be delayed (assumed to be an integer). This delay is the sum of two attributes: the *Departure delay* and the *Sailing delay* (both of which are expressed as an integer number of days). Before the ship leaves port it could be delayed for *Loading* problems; a *Forecast* of foul weather could cause the Captain to delay departure; and *Maintenance* could cause the ship to sit at the dock (we simplify these to true/false variables for the example). For simplicity, we assume that each of these factors delay departure by one day. Therefore the total *Departure delay* could be up to three days. Similarly, bad *Weather* en route could cause delays, as could need making *Repairs* at sea (again simplified to true/false variables). These delays contribute to the *Sailing delays*, again an integer number of days. Figure 1 shows the evidential network for the problem.

In the network, there are 8 variables represented by the circles and 7 valuation variables represented by the diamond-shaped rectangles. The belief functions representing the relations among the variables are defined on the valuation variables connecting the variables they include, the details about the relations among the variables are shown in the appendix. We can also provide prior beliefs for some variables, regarded as the evidence, which are stored in the valuation variables connected to those variables. In this example,

---

[2]The example is abstracted from [1] with minor changes on some prior beliefs.

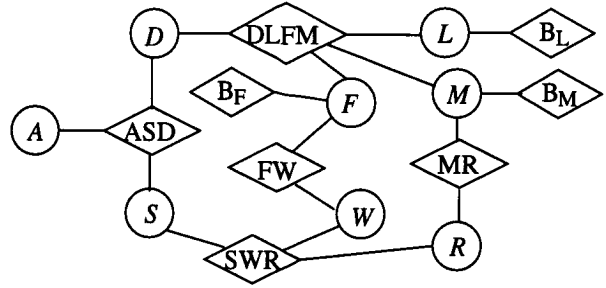

Figure 1: An evidential network for the Captain problem where $\mathbf{U}=\{A, D, S, F, L, M, W, R\}$.

we have three pieces of evidence on the variable $L$, $M$ and $F$, respectively, shown in table 1.

| Variable | $m(\{T\})$ | $m(\{F\})$ | $m(\Theta)$ |
|---|---|---|---|
| $L$ | .3 | .50 | .20 |
| $F$ | .2 | .60 | .20 |
| $M$ | 0 | .95 | .05 |

Table 1: The prior belief functions for $L$, $F$, and $M$.

As the goal of the example is to find how many days late the ship might arrive in port, the marginal for the variable $A$ will be the focus of our attention. This can be obtained by evaluating the network using local computation. Note that the possible value of $A$ is $\Theta_A=\{0,1,2,3,4,5,6\}$. The result is shown in table 2.

| focal element $a$ | $m(a)$ | $a$ | $m(a)$ |
|---|---|---|---|
| $\{0,1,2,3,4,5,6\}$ | .0002 | $\{2,3,5\}$ | .00005 |
| $\{0,1,2,3,4,5\}$ | .0053 | $\{3,4\}$ | .0055 |
| $\{1,2,3,4,5,6\}$ | .0005 | $\{2,3\}$ | .0696 |
| $\{0,1,2,3,4\}$ | .0452 | $\{1,2\}$ | .1145 |
| $\{1,2,3,4,5\}$ | .0117 | $\{0,1\}$ | .1564 |
| $\{2,3,4,5,6\}$ | .0003 | $\{1,3\}$ | .0137 |
| $\{1,2,3,4\}$ | .0712 | $\{0,2\}$ | .0139 |
| $\{0,1,2,3\}$ | .1038 | $\{2,4\}$ | .0086 |
| $\{2,3,4,5\}$ | .0070 | $\{4,5\}$ | .0001 |
| $\{1,2,3,5\}$ | .00005 | $\{0\}$ | .0410 |
| $\{2\ 3\ 4\}$ | .0565 | $\{1\}$ | .0681 |
| $\{1,2,3\}$ | .0521 | $\{2\}$ | .0392 |
| $\{3,4,5\}$ | .00003 | $\{3\}$ | .0137 |
| $\{0,1,2\}$ | .0933 | $\{4\}$ | .0085 |
| $\{0,2,4\}$ | .00007 | $\sum m(a)$ | 1.0000 |

Table 2: Marginal of global belief function for *Arrival delay*.

Generally, it is difficult to interpret the raw focal elements for the non-binary variables. So we will look at beliefs and plausibilities for the singleton subsets of $\Theta_A$ which correspond to each day and for the singleton subsets of a coarsening frame $\Theta'_A = \{\{0,1\}, \{2,3\}, \{4,5,6\}\}$. Table 3 shows the result.

| a | bel(a) | pl(a) |
|---|--------|-------|
| {0} | .0410 | .4592 |
| {1} | .0681 | .7360 |
| {2} | .0392 | .6929 |
| {3} | .0137 | .4563 |
| {4} | .0085 | .2207 |
| {5} | 0 | .0252 |
| {6} | 0 | .0010 |

| a | bel(a) | pl(a) |
|---|--------|-------|
| {0,1} | .2656 | .7910 |
| {2,3} | .1225 | .7258 |
| {4,5,6} | .0086 | .2208 |

Table 3: Beliefs and plausibilities for some subsets of *Arrival delay*.

Considering frame $\Theta_A$, we find that the most plausible day that the ship might delay is one day, and that it has the strongest support. However, the belief on the subset $\{1\}$ is very small. So we look at a coarsening frame $\Theta'_A$. From $\Theta'_A$, we find that being late within one day is strongest supported and more than 4 days is hardly plausible. Now we would like to know the origin of the support given to the conclusion. Since there are several prior beliefs (or pieces of evidence), which one is most important to the result? In the rest of this section, we will discuss two strategies to answer such kinds of questions.

### 3.1 Sensitivity of the support for the hypotheses of a piece of evidence

Consider $n$ distinct pieces of evidence $\mathcal{E}_i$ (i=1, ..., ,n) on $\Theta$ represented by belief functions $bel_i$. $bel$ quantifies the combined impact of the n pieces of evidence. Suppose $x \in \Theta$, $bel(\{x\}) \geq bel(\{z\})$ for all $z \in \Theta$, $z \neq x$. Strat [?] has proposed a tool to explain why a particular hypothesis was found to be strongly (weakly) supported based on sensitivity analysis. Since the requirement of systematic variation for sensitivity analysis is not feasible for the case of belief functions, Strat proposed to use discounting operation. The idea is to first use the discounting operation for each evidence. i.e.,

$$m_i^{disc}(a) = \begin{cases} \alpha_i m_i(a) & a \neq \Theta \\ 1 - \alpha_i + \alpha_i m_i(\Theta) & \text{otherwise.} \end{cases}$$

where $\alpha_i$ is the credibility of the original evidence $\mathcal{E}_i$, and then proceed the following,

1. compute for each evidence:
$$\widehat{bel}_i(x) = \left. \frac{\partial bel^{disc}(x)}{\partial \alpha_i} \right|_{\alpha_j=1}$$
$$\widehat{pl}_i(x) = \left. \frac{\partial pl^{disc}(x)}{\partial \alpha_i} \right|_{\alpha_j=1}$$
where j=1, ..., n. Here $\widehat{bel}_i(x)$ can be interpreted as the sensitivity of the support for x of $\mathcal{E}_i$, and likewise for $\widehat{pl}_i(x)$.

2. Identify those $\mathcal{E}_i$ with the extreme values.

In general, the positive values of $\widehat{bel}_i(x)$ and $\widehat{pl}_i(x)$ indicates the support to the conclusion, while the negative one indicates that the evidence argues against the conclusion. The larger the absolute value of $\widehat{bel}_i(x)$ or $\widehat{pl}_i(x)$ is, the greater the impact of $\mathcal{E}_i$ upon the hypothesis is. Positive $\widehat{bel}_i(x)$ and negative $\widehat{pl}_i(x)$ means decreasing the ignorance without necessarily arguing for or against hypothesis, while negative $\widehat{bel}_i(x)$ and positive $\widehat{pl}_i(x)$ indicates adding to the confusion about the hypothesis. Note that in [?], the normalized belief function is used for the analysis. In this paper, we will always use the unnormalized belief. Moreover, instead of analyzing the impact on the belief of a single hypothesis, we also consider the case for some subsets of the hypotheses if they are meaningful.

Strat [?] showed that, in practice, numeric techniques are required to compute these quantities, and in reality, they are computed by:

$$\widehat{bel}_i(x) \approx \frac{[bel_i^{disc}(x)]_{\alpha_i=1} - [bel_i^{disc}(x)]_{\alpha_i=1-\delta}}{\delta}$$

for some small $\delta$. So as for $\widehat{pl}_i(x)$. The following theorem [17] shows that if we do not consider the normalization factor, $\widehat{bel}_i(x)$ and $\widehat{pl}_i(x)$ are constants, and thus can be computed precisely.

**Theorem 1** *Consider n distinct pieces of evidence $\mathcal{E}_i$ (i=1, ..., n) on $\Theta$ represented by $bel_i$. Let bel quantify the combination of the n pieces of evidence, $bel^{-i}$ be the combination of n-1 belief functions except $bel_i$, i.e., $bel^{-i} = \oplus\{bel_j | j = 1, \ldots, n, j \neq i\}$. Then, $\forall x \subseteq \Theta$*

$$\widehat{bel}_i(x) = bel(x) - bel^{-i}(x),$$
$$\widehat{pl}_i(x) = pl(x) - pl^{-i}(x).$$

From theorem 1, we find that $\widehat{bel}_i(x)$ is in fact the difference of $bel$ for $x$ between the cases when $\mathcal{E}_i$ is considered and not considered given the other evidence. So as for $\widehat{pl}_i(x)$. It is not difficult to derive that the similar result holds in the evidential systems.

**Corollary 1** *Let $\mathbf{U} = \{X_1, \ldots, X_n\}$ be the set of the variables in an evidential system. Suppose $X_j (\in \mathbf{U})$ is the hypothesis variable that we are interested in, and suppose we have prior beliefs on some variables. Then the sensitivity of the impact of variable $X_i$ on $x \subseteq \Theta_{X_j}$ is computed by:*

$$\widehat{bel}_i(x) = (\oplus\{bel_A | A \in \mathbf{H}\})^{\downarrow X_j}(x)$$
$$- (\oplus\{bel_A | A \in (\mathbf{H} - \{X_i\})\})^{\downarrow X_j}(x), \quad (1)$$

*and likewise for $\widehat{pl}_i(x)$.*

From eq.(1), we find that, to compute $\widehat{bel}_i(x)$, we only need to compute the differences of the *bel* and *pl* for $x$ between the cases where $\mathcal{E}_i$ is considered and not considered. This makes the computation easier.

**Example** (continued): According to theorem 1, we compute the sensitivity of the support to the hypotheses of $L$, $F$, and $M$ respectively. The result is shown in table 4.

| $a$ | $\widehat{bel}$ | $\widehat{pl}$ | $\widehat{bel}$ | $\widehat{pl}$ | $\widehat{bel}$ | $\widehat{pl}$ |
|---|---|---|---|---|---|---|
|  | \multicolumn{2}{c}{$L$} | \multicolumn{2}{c}{$F$} | \multicolumn{2}{c}{$M$} |
| {1} | .068 | -.113 | .068 | -.141 | .020 | -.049 |
| {6} | 0 | -.001 | 0 | -.002 | 0 | -.001 |
| {0,1} | .181 | -.058 | .266 | -.155 | .084 | 0 |
| {4,5,6} | .009 | -.172 | .009 | -.251 | 0 | -.129 |

Table 4: Sensitivity of support to the hypotheses of each piece of evidence.

Let's first look at the change of the belief and of the plausibility for the singleton subsets of $\Theta_A$. From table 4, it can be found that none of the three pieces of evidence explicitly argue for or against the hypothesis "being one day late" since $\widehat{bel}_X(a) > 0$ and $\widehat{pl}_X(a) < 0$ ($X \in \{F, M, L\}$). All three ague against {6} since $\widehat{bel}_X(a) = 0$ and $\widehat{pl}_X(a) < 0$ and the evidence on *Forcast* has the largest impact. Now consider the support for {0, 1} and {4, 5, 6}, we have that the evidence on *Maintenance* is the only one supporting {0, 1} and arguing against {4, 5, 6} while the other two only decrease the ignorance.

### 3.2    Analysis of the Measure of Information provided by a piece of evidence

Apart from sensitivity analysis, another way to explain the reasoning process is to analyze the amount of information provided by the evidence, thus the impact of the evidence on the overall conclusion instead of on a single hypothesis. Measures of information are often quantified such that the additivity property holds. In the theory of belief functions, Smets [8] gave the following definition:

**Definition 2** *Let $I(bel)$ denote the amount of information in a piece of evidence $\mathcal{E}$ represented by a belief function bel on $\Omega$. Then $I(bel)$ is computed by:*

$$I(bel) = -\sum_{a \subseteq \Omega} \log q(a)$$

*where $q$ is the commonality function.*

Note that in this paper all the beliefs should be such as $m(\Omega) > 0$, called the non-dogmatic belief functions. In this case, $I(bel)$ is non-negative. Otherwise $I(bel)$ is infinite, for which case Smets gave a discussion in [8]. From definition 2, we find that a vacuous belief function contains no information, i.e., $I(bel)=0$. The following lemma states the additivity property such that the amount of information of the combination of two distinct pieces of evidence[3] is the sum of the information of these two pieces of evidence.

**Lemma 1** *Consider two distinct pieces of evidence $\mathcal{E}_1$ and $\mathcal{E}_2$ on $\Omega$ represented by $bel_1$ and $bel_2$. $bel_{12}$ quantifies the combined impact of $\mathcal{E}_1$ and $\mathcal{E}_2$. Then*

$$I(bel_{12}) = I(bel_1) + I(bel_2).$$

It is easy to generalize lemma 1 to the case of $n$ distinct pieces of evidence: Let $bel$ denote the belief quantifying the combined impact of the $n$ pieces of evidence. We have:

$$I(bel) = \sum_{i=1}^{n} I(bel_i). \qquad (2)$$

Therefore, we define the information brought by a distinct piece of evidence $\mathcal{E}_{n+1}$ as following:

$$\Delta I^{n+1} = I(bel \oplus bel_{n+1}) - I(bel) = I(bel_{n+1}). \qquad (3)$$

Then the explanation is as follows: Let $\mathcal{E}_1, \ldots, \mathcal{E}_n$ be $n$ distinct pieces of the evidence on $\Omega$. $\mathcal{E}_i$ brings the most information on $\Omega$ or $\mathcal{E}_i$ is the most important evidence iff $\Delta I^i \geq \Delta I^j$ or $I(bel_i) \geq I(bel_j)$ for $j = 1, \ldots, n$.

Generally, in an evidential network, let $X$ be a hypothesis variable that we are interested in, $X_1, \ldots, X_n$ be some evidence variables. Suppose we have prior beliefs $bel_{0X_i}$ on some of $X_i$'s. After propagation, we can get the marginal of the global belief function $bel_0$ for $X$. In the rest of this section, we analyze how much information each piece of evidence has brought to $X$, thus explain which piece of evidence is most important to the conclusion.

**Definition 3** *In an evidential network, suppose $X$ is a hypothesis variable that we are interested in. Let $bel_0$ be the global belief function where all the prior beliefs are vacuous Suppose there is one prior belief $bel_{0X_i}$ on some variable $X_i$. Then we define the amount of information that $X_i$ has brought to $X$ individually as:*

$$I(X_i) = I((bel_0 \oplus bel_{0X_i})^{\downarrow X}) - I(bel_0^{\downarrow X}),$$

*Suppose $A \subseteq \mathbf{U}$, then the amount of information that $A$ has brought to $X$ is defined as:*

$$I(A) = I((bel_0 \oplus \{bel_{0X_i} | X_i \in A\})^{\downarrow X}) - I(bel_0^{\downarrow X}).$$

**Definition 4** *Let $bel^{-i} = bel_0 \oplus \{bel_{0X_j} | X_j \neq X_i\}$. Then we can define the amount of information that $X_i$ brought to $X$ given the other evidence as:*

$$\Delta I(X_i) = I(bel^{\downarrow X}) - I((bel^{-i})^{\downarrow X}), \qquad (4)$$

---

[3]Smets [10] has given a definition for the concept of distinct evidence

We can also define $\Delta I(B)$ for a subset $B$: let $bel^{-B} = bel_0 \oplus \{bel_{0X_j} | X_j \notin B\}$, then we have:

$$\Delta I(B) = I(bel^{\downarrow X}) - I((bel^{-B})^{\downarrow X}).$$

Generally, the amount of information that a subset $A$ brought to $X$ is not the sum of the information brought to $X$ individually, i.e., $I(A) \neq \sum_{X_i \in A} I(X_i)$. When there are more than one pieces of evidence, the amount of information that $X_i$ brought to $X$ given the other evidence does not always equal to the information it brought to $X$ individually, i.e., $\Delta I(X_i) \neq I(X_i)$. This is because the combination and coarsening are not commutative. However, the conclusion is different when there exist the relations of conditional independence among the variables. Theorem 2 and its corollary will illustrate such relations. First, let's look at the concept of conditional independence in VBS, which is given in [7].

**Definition 5** *Suppose $S$, $T$, and $X$ are disjoint subsets of $\mathbf{U}$. We say $S$ and $T$ are conditionally independent given $X$, written as $S \perp T \mid X$, if and only if $bel^{\downarrow X \cup S \cup T} = bel^{\downarrow X \cup S} \oplus bel^{\downarrow X \cup T}$.*

**Theorem 2** *Let $A$, $B$ be two disjoint subsets of $\mathbf{U}$, $X$ be the hypothesis variable we are interested in, and $A \perp B \mid X$[4]. Suppose $bel_0^{\downarrow X}$ is vacuous and only the variables in $A$ or $B$ have prior beliefs. Then we have the following relations for the information measures. (Proof can be found in [17])*

$$I(A \cup B) = I(A) + I(B), \quad \Delta I(A \cup B) = \Delta I(A) + \Delta I(B),$$
$$\Delta I(A) = I(A), \quad \Delta I(B) = I(B)$$

**Corollary 2** *Suppose all the variables which have prior beliefs are conditionally independent given $X$, i.e., for any $X_i$, $X_j$ such that $bel_{0X_i}$, $bel_{0X_j}$ are not vacuous, $X_i \perp X_j \mid X$. And suppose $bel_0^{\downarrow X}$ is vacuous. Then $\Delta I(X_i) = I(X_i)$, and $I(A) = \sum_{X_i \in A} I(X_i)$ where $A$ is the set of all the variables with prior beliefs.*

In an evidential system, when we analyze the impact of a piece of evidence on the conclusion, it generally depends on the situation where the other pieces of evidence are given. Thus, to answer the question such as: which evidence brings most (less) information or which evidence is the most (less) important to the conclusion, we select the one whose $\Delta I(X_i)$ is the biggest.

**Example** (continued): By applying eq. (4) in the captain example, we have:

$$\Delta I(L) = I(bel^{\downarrow A}) - I((bel^{-L})^{\downarrow A}) = 60.47$$
$$\Delta I(F) = I(bel^{\downarrow A}) - I((bel^{-F})^{\downarrow A}) = 67.60$$
$$\Delta I(M) = I(bel^{\downarrow A}) - I((bel^{-M})^{\downarrow A}) = 102.95.$$

---
[4]We write $X$ for $\{X\}$ when confusion is absent.

Therefore, for the question "which evidence is the most important to the conclusion?", the evidence on <u>M</u>aintenance would be its answer.

## 4 The Implementation Issue

We have presented the approaches for the explanation. Now we discuss their implementation. First, we introduce the concept of removal. Smets [14] defined the operator $\ominus$ as the inverse of the operator $\oplus$ in the sense that:

$bel_1 \oplus bel_2 \ominus bel_2 = bel_1$ for any $bel_1$ and $bel_2$ on $\Omega$, and $bel_1 \ominus bel_1$ is a vacuous belief function.

In VBS, Shenoy [7] gave the definition of *removal* $\ominus$ for belief functions as follows: Considering two valuations represented by the commonality functions $q_A$ and $q_B$ on the subsets $A$ and $B$. Then $q_A \ominus q_B$ will be on $A \cup B$ computed by[5]: $\forall c \subseteq \Theta_{A \cup B}$,

$$(q_A \ominus q_B)(c) = \frac{q_A(c^{\downarrow A})}{q_B(c^{\downarrow B})}.$$

Given the operator $\ominus$, eq. (1) for the computation of the sensitivity of support can be rewritten as:

$$\widehat{bel}_i(x) = bel^{\downarrow X}(x) - (bel \ominus bel_{0X_i})^{\downarrow X}(x),$$
$$\Delta I(X_i) = I(bel^{\downarrow X}) - I((bel \ominus bel_{0X_i})^{\downarrow X}),$$

where $bel$ is the global belief function and $X$ is the variable we are interested in.

As mentioned before, we use local computational technique to compute $bel^{\downarrow X}$. To compute $(bel \ominus bel_{0X_i})^{\downarrow X}$, one way is to remove the prior belief for each variable and repropagating the changes in the whole network each time. This needs a lot of redundant computation. From the definition of removal operator, it is easy to have:

$$(bel \ominus bel_{0X_i})^{\downarrow X} = (bel^{\downarrow(X \cup X_i)} \ominus bel_{0X_i})^{\downarrow X} \quad (5)$$

That's to say, if we compute $bel^{\downarrow(X \cup X_i)}$ by local computation, then we can compute $(bel \ominus bel_{0X_i})^{\downarrow X}$ directly, Xu [18] has proposed a method to compute the marginals for any subsets from the marginal representation, avoiding the unnecessary computation. Therefore, $(bel \ominus bel_{0A_i})^{\downarrow X}$ can be computed as follows:

1. From the marginal we have computed, we first compute the marginals $bel^{\downarrow X \cup X_i}$ for all $X_i'$s which have prior beliefs. This can be done in parallel according to [18];

---
[5]The removal operation for the case of belief function may result in a non-belief function if $q_B$ has not been combined in $q_A$ before being removed. In this paper, this will not happen since in the later discussion, all the valuations to be removed are those having been combined previously.

2. Compute $(bel \ominus bel_{0X_i})^{\downarrow X}$ by applying eq. (5).

Then we can easily compute $\widehat{bel}_i(x), \widehat{pl}_i(x)$ and $\Delta I(X_i)$ for the explanation purpose.

## 5 Conclusions

We have presented two approaches to explain reasoning process in general evidential systems. One is to compute the sensitivity of the support to a hypothesis based on sensitivity analysis, the other is to explain which evidence is the most important to the whole conclusion based on the measure of information that the evidence provides. We have also discussed the possibility of implementing these approaches. In this paper, we only consider the impact of a single piece of evidence. However, the approaches can be generalized to find the strongest relevant subset of factual knowledge (evidence) $R_{fact}$ to the conclusion by a stepwise procedure as following:

1. Find the strongest relevant evidence (whose has greatest impact on the conclusion) $X_i$ such that $\Delta I(X_i) \geq \Delta I(X_k)$. Let $R_{fact} = \{X_i\}$;

2. Find the next strongest one $X_j$ such that $\Delta I(R_{fact} \cup \{X_j\}) \geq \Delta I(R_{fact} \cup \{X_k\})$, $X_j, X_k \notin R_{fact}$. Let $R_{fact} = R_{fact} \cup \{X_j\}$;

3. Do step 2 iteratively until the increase in $\Delta I(R_{fact} \cup \{X_j\})$ is negligible when adding more $X_j$ in $R_{fact}$. Then we find a small set of strong relevant evidence. e.g., in figure 2, when the number of elements in $R_{fact}$ is larger than 6, the increase in $\Delta I(R_{fact})$ can be negligible, thus we keep the first 6 pieces of evidence in $R_{fact}$.

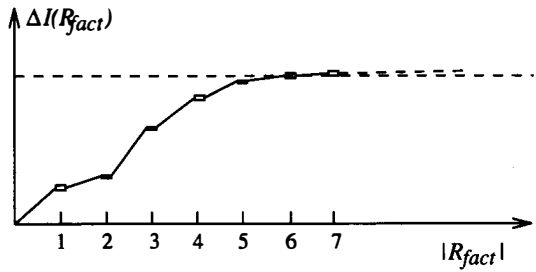

Figure 2: the information brought by the first n strongest relevant pieces of evidence.

4. Since $\Delta I(X_i)$ is the amount of information that $X_i$ brought given the other evidence, it is possible that $\exists X_i \in R_{fact}$ such that $\Delta I(X_i) > \Delta I(X_k), X_k \in R_{fact}$, but $\Delta I(X_i) \ll \Delta I(R_{fact} - \{X_i\})$. Then $X_i$ can not be regarded as a strong relevant piece of evidence to the conclusion since its impact depends on the appearance of the other evidence and thus should be removed from $R_{fact}$. After all such kind of $X_i$ removed, the rest $R_{fact}$ will be the strongest relevant subset of factual knowledge.

In this paper, we focus our attention on the numeric computation, further research on integrating the qualitative analysis and on the natural language processing is needed to perfect the explanation facilities and to provide a better user-friendly interface.

## Acknowledgment

The authors would like to thank the anonymous referees for their comments. This research work has been partially supported by the ESPRIT III, Basic Research Project Action 6156 (DRUMS II) funded by a grant from the Commission of the European Communities. The first author is supported by a grant of IRIDIA, Université libre de Bruxelles.

## Appendix: Belief Functions on the Valuation Variables for the Example:

The relationship among $A$, $D$, $S$ can be expressed as $a = d + s$ where $a \in \Theta_A = \{0...6\}$, $d \in \Theta_D = \{0...3\}$ and $s \in \Theta_S = \{0...3\}$. Then belief function for $\{A, D, S\}$ is:

$$m\left(\left\{\begin{array}{cccc} (0,0,0) & (1,0,1) & (2,0,2) & (3,0,3) \\ (1,1,0) & (2,1,1) & (3,1,2) & (4,1,3) \\ (2,2,0) & (3,2,1) & (4,2,2) & (5,2,3) \\ (3,3,0) & (4,3,1) & (5,3,2) & (6,3,3) \end{array}\right\}\right) = 1.$$

For $\{D, L, M, F\}$, any of $L$, $M$, or $F$ being true adds one day to the total delay $D$. The belief function is:

$$m\left(\left\{\begin{array}{ccc} (0,F,F,F) & (1,T,F,F) & (1,F,T,F) \\ (1,F,F,T) & (2,F,T,T) & (2,T,F,T) \\ (2,T,T,F) & (3,T,T,T) & \end{array}\right\}\right) = 1.$$

Similar for the $\{S, W, R\}$, but we expect our rule to be accurate 90% of time. Then,

$$m(\{(0,F,F),(1,F,T),(1,T,F),(2,T,T)\}) = .9$$
$$m(\Theta_S \times \Theta_W \times \Theta_R) = .1.$$

If the weatherman was 100% accurate, we could express the rule $\{F, W\}$ as $F \leftrightarrow W$. But we have only 80% confidence in the weatherman. Thus:

$$m(\{(F,F),(T,T)\}) = .8$$
$$m(\Theta_W \times \Theta_R) = .2.$$

If we have a separate body of evidence that indicates that the ship breaks down between 20 and 80% of the time after no maintenance and that 10 to 30% after maintenance. $R$ given $M$:

| belief of $R$ given $M$ | T | F |
|---|---|---|
| $\{T\}$ | .1 | .2 |
| $\{F\}$ | .7 | .2 |
| $\Theta_R$ | .2 | .6 |

Belief on the product space $R \times M$ can be computed by using so-called ballooning extension proposed by Smets [12]:

$$m(\{(T,T),(T,F),(F,F)\}) = .06$$
$$m(\{(T,T),(T,F)\}) = .02$$
$$m(\{(T,T),(F,F)\}) = .02$$
$$m(\{(T,F),(F,T),(F,F)\}) = .42$$
$$m(\{(T,F),(F,T)\}) = .14$$
$$m(\{(F,T),(F,F)\}) = .14$$
$$m(\{(T,T),(T,F),(F,T)\}) = .04$$
$$m(\{(T,T),(F,T),(F,F)\}) = .04$$
$$m(\Theta_R \times \Theta_M) = .12.$$